\documentclass{article}

\PassOptionsToPackage{numbers,compress}{natbib}

\usepackage[dblblindworkshop,final]{neurips_2026}
\usepackage[utf8]{inputenc}
\usepackage[T1]{fontenc}
\usepackage{hyperref}
\usepackage{url}
\usepackage{booktabs}
\usepackage{amsfonts}
\usepackage{amsmath}
\usepackage{nicefrac}
\usepackage{microtype}
\usepackage{xcolor}
\usepackage{graphicx}
\usepackage{array}
\graphicspath{{figs/}}

\title{Attributing Agentic Failure in Reasoning Vision-Language Models with Interactive 2D Environments}
\workshoptitle{VLM4RWD}

\author{%
    Pranav Guruprasad$^{1,2}$ \quad Sean Rivera$^{2}$ \quad Helen Lu$^{2,3}$ \quad Arushi Jain$^{2}$ \\
    \textbf{Hangliang Ren}$^{2}$ \quad \textbf{Harshvardhan Sikka}$^{1,2,4}$ \\
    $^{1}$Fig \quad $^{2}$Manifold Research Group \quad $^{3}$Tufts University \quad $^{4}$Georgia Institute of Technology \\
    \texttt{pranav@fig.inc} \\
  } 

\begin{document}

\maketitle

\begin{abstract}
  Agentic benchmarks report a success rate that confounds distinct causes of
  failure: misreading the environment, never working out how components of the
  environment operate, operating the components in the wrong order, and failing
  to recover from a mistake, each of which calls for a different fix. We present an early preview of MultiNet v2.0 -
  an interactive maze environment built for attribution rather than ranking. Four
  different types of mechanisms, namely key-door pairs, switch-gate pairs,
  distractors and decoys, and ordered combinations of these mechanisms, can be
  added, modified, or removed independently while the renderer, the action space,
  and the scoring function stay fixed. A breadth-first search over the joint state
  of position, facing direction, inventory, and mechanism state gives every
  instance an exact optimal action count, which sets its difficulty and grades
  unfinished episodes. We fix the evaluation protocol on 15 held-out mazes over
  540 episodes, then evaluate five vision-language models on 50 mazes. Three
  models solve 6 of 150 episodes, and two newer frontier models, GPT-6 Astra and
  Claude Opus 5.5, solve 18 of 100. Across all
  five models, 38 of the 50 mazes were not solved, and no episode containing a
  switch-gate mechanism was solved.
\end{abstract}

\section{Introduction}

Language and vision-language models are increasingly deployed to act over many
steps in environments whose rules they are never given, including browsers,
operating systems, code repositories, and robots
\citep{zhou2023,xie2024,jimenez2023,guruprasad2024benchmarking,guruprasad2025generality}. Evaluations that report only whether such
an agent finished a task leave the practitioner without a way to choose an
intervention. An agent that never worked out what a switch does needs a
different fix from an agent that understood every mechanism but operated them
out of order, and both differ from an agent that knew what to do and issued
actions that did not achieve it. A high aggregate score can hide such a failure entirely: GUI grounding models that exceed 85\% on standard benchmarks lose 27 to 56 percentage points once instructions require spatial reasoning, and fine-tuning on augmented data, the obvious remedy, lowers their accuracy rather than restoring it \citep{wang2026guiperturbed}. A survey of agentic benchmark construction also found task setup and reward design flaws that change reported performance by up to 100\% in relative terms \citep{zhu2025}, so pursuing the wrong explanation for a failure is a real cost.

We therefore ask a narrower question than whether current models can act over
long horizons. We ask where and how they break when the rules of the
environment have not been explained to them. Our environment is a
two-dimensional maze rendered with a top-down view, in which an agent starts at a fixed cell
facing a fixed direction and must reach a goal using an action space containing six different actions. It is
told the action space, the goal, the inventory it contains at the current timestep, text summary of its previous actions, and nothing else as language input, so what a key does, what a
switch does, and which door a given key opens must be inferred from how the
state changes in response to actions.

Separating causes of failure places four requirements on an environment, and
the maze substrate satisfies all four at once. The environment must be
\textbf{knowledge-free}, so that failure is not an artifact of an interface the model has never seen \citep{wang2026guiperturbed} or, conversely, of an interface it has memorized \citep{chollet2019}. It must be \textbf{generative}, so that difficulty scales
with model capability rather than saturating against a fixed answer key
\citep{kiela2021,cobbe2020}. It must be \textbf{projectable}, so that the same
task can be rebuilt in a different modality such as text or three-dimensional
simulation, which tests whether a model understands the task rather than one
rendering of it, since performance can shift with the input type alone \citep{guruprasad2025generality}. It must be \textbf{verifiable}, so that success, failure, and
partial progress can be measured against an optimal solution
\citep{anderson2018}.

Verifiability is what makes attribution quantitative here. Because a maze is a
finite deterministic system, a breadth-first search over the joint state of
position, facing direction, inventory, and mechanism state returns the exact
optimal action sequence from any reachable state. Every instance therefore
carries an objective difficulty in executable actions, and every unfinished
episode receives graded credit that reflects mechanism progress rather than
proximity to a goal that may be sealed behind an unopened door.

Three vision-language models, Claude Opus 4.8, Kimi K2.6, and Qwen 3.6-27B, solve 6 of 150 episodes on our 50-maze evaluation set, and 45 of the 50 mazes are solved by none of them. The failures separate cleanly: keys and doors are handled with partial success, no episode containing a switch and gate was solved, and each model has a distinct terminal behavior. Two newer frontier models, GPT-6 Astra and Claude Opus 5.5, solve 18 of 100 episodes and their success depends far less on path length compared to the other evaluated models, and yet the same attribution holds: neither solves any of the 35 mazes containing a switch and gate, and each again degrades in its own way.

As a part of this early preview of MultiNet v2.0, we contribute an interactive environment in which four mechanism families can be added, modified, or removed independently against a fixed renderer, action space, and scoring function (Section~\ref{sec:env}); an exact oracle over the joint environment state that supplies per-instance difficulty in executable actions and a graded progress metric crediting subgoals (Section~\ref{sec:scoring}); a 540-episode protocol sweep over 12 conditions on 15 held-out mazes, which fixes the evaluation protocol before any benchmark number is measured
(Section~\ref{sec:protocol}); an evaluation of three vision-language models over 150 episodes attributing failure to path length, mechanism familiarity, and per-model terminal behaviors (Section~\ref{sec:results}); an evaluation of two newer frontier models showing that the switch-gate failure persists as overall performance rises (Section~\ref{sec:frontier}); and an open-source release\footnote{\label{fn:code}Code: \url{https://github.com/ManifoldRG/MultiNet-v2.0}. Project page: \url{https://multinet.ai}.}
of the environment, harness, run infrastructure, and oracle as part of this early preview of MultiNet v2.0   (Appendix~\ref{app:release}).

\section{Related work}

\paragraph{Agentic benchmarks in realistic environments.}
SWE-bench \citep{jimenez2023}, WebArena \citep{zhou2023}, OSWorld
\citep{xie2024}, ALFWorld \citep{shridhar2021}, MultiNet v1.0 \citep{guruprasad2025generality}, and $\tau$-bench
\citep{yao2024} evaluate agents in software repositories, browsers, desktops,
household simulators, offline robotics setups, multi-turn conversational function calling settings, and tool-mediated dialogue. Two properties limit their
diagnostic use. Because they reproduce real interfaces, a model has typically
seen GitLab, an e-commerce checkout, or a Linux file manager during pretraining,
so a success does not distinguish recall of a familiar interface from
acquisition of a new affordance \citep{chollet2019}. And because they score
functional correctness with a programmatic check on a terminal state, they
report whether an agent succeeded but not how far it advanced, and supply no
per-instance measure of difficulty. We give up realism in order to control both properties. 

\paragraph{Games and procedurally generated environments.}
BALROG \citep{paglieri2024} evaluates agents zero-shot across BabyAI
\citep{chevalier2019}, Crafter \citep{hafner2022}, TextWorld \citep{cote2018},
MiniHack \citep{samvelyan2021}, Baba is AI \citep{cloos2024baba}, and NetHack \citep{kuttler2020} under both
textual and visual observations. Our finding that visual observation is harder
agrees with theirs. Our earlier work MultiNet v0.2 \citep{guruprasad2025procedural} evaluated vision-language and vision-language-action models on the procedurally generated games of Procgen \citep{cobbe2020}. The designs differ in what is varied. BALROG varies the
environment itself, aggregating six games with different rules, renderers, and
action spaces into one score, so a gap between two games is not attributable to
any single cause. We vary factors inside one environment, holding renderer,
action space, and scoring fixed while adding, modifying, or removing a single mechanism. We
build directly on MiniGrid \citep{chevalier2023}.

\paragraph{Interactive reasoning with controlled priors.}
ARC-AGI-3 \citep{arcprize2025,arcprize2026} is an interactive benchmark of
turn-based environments presented without instructions, rules, or stated goals,
on which human testers solve every environment while frontier systems perform poorly. Our setting shares its commitments to unexplained
interactive puzzles with priors controlled by design. We differ in two respects.
ARC-AGI-3 environments are handcrafted, so two may differ along many dimensions
at once, whereas our instances are compositions of a declared parameter set, so
the difference between any two mazes is known and their difficulties are
comparable. Second, the maze substrate admits an exact optimal action sequence
from any reachable state, which converts a binary outcome into graded credit.

\paragraph{Scoring partial progress.}
\citet{anderson2018} set out evaluation measures for embodied navigation that
weight success against the length of the shortest path. Our progress metric
extends that idea to a state space that includes mechanism state, which matters
because an agent can stand adjacent to a goal that is unreachable until a door
is opened. PlanBench \citep{valmeekam2023} grades plans against optimal
solutions in symbolic domains, whereas our oracle grades an interleaved sequence
of perception, inference, and execution in a closed loop.

\section{The environment}
\label{sec:env}

%

\subsection{Task and action space}

An instance is a maze on a grid of between $8 \times 8$ and $14 \times 14$
cells. The agent begins at a fixed start cell facing a fixed direction and must
reach a goal tile. Six actions are available: turn left, turn right, move
forward, pickup, toggle, and done. Movement and turning are egocentric, so the
effect of an action depends on the direction the agent currently faces. The
agent is told the action space and the goal, and nothing about what any object
is for, which makes the task one of affordance discovery in the sense of
\citet{gibson1979} rather than instruction following.

\subsection{Mechanisms and what each one isolates}

Solving an instance requires four things: working out how each mechanism
operates, reasoning about the order in which mechanisms must be operated,
executing the intended actions correctly, and recovering when an attempt fails.
Each mechanism family is present because it isolates one of these four
requirements, and each can be added or removed without altering the others (Appendix \ref{app:mechanisms}, Table \ref{tab:mechanisms}).

A key is acquired by issuing a pickup action while standing on its cell, which adds it to
inventory. A door is opened by issuing toggle from an adjacent cell, and opens
only if the matching key is held. A switch is flipped by issuing toggle while
standing on its cell, and its gate opens immediately. The two mechanisms are
deliberately close in interaction cost and far apart in prior familiarity, which
is what allows the comparison in Section~\ref{sec:failure} to separate discovery
from execution. A dependency chain composes mechanisms so that ordering is forced,
for example a gate whose switch sits behind a door that requires a key.

\subsection{Observations}

At each turn the model receives its start position, a prose summary listing
every mechanism event so far in order, and up to four rendered frames: the three
most recent steps, each labelled with the inventory and the action chosen at
that step, followed by the current frame. Frames are top-down images of the
whole maze, so the environment is fully observed visually and partially observed
with respect to mechanism semantics \citep{kaelbling1998}, which stay hidden
until an interaction reveals them. The model receives no textual statement of
how the environment changed in response to its last action, so that change must
be inferred by comparing consecutive frames. Section~\ref{sec:protocol} explains
why this is a deliberate difficulty setting.

\subsection{Scoring and the oracle}
\label{sec:scoring}

An episode is a \emph{solve} when the agent reaches the goal tile. Because
solves are rare, we report a graded progress score as the primary continuous
measure. Let $d(s)$ be the remaining oracle cost from state $s$, let $s_0$ be
the start state, and let $s_T$ be the state at which the episode terminates.
Progress is

\begin{equation}
  p = 1 - \frac{d(s_T)}{d(s_0)} .
  \label{eq:progress}
\end{equation}

Progress equals one at a solve and is at most zero when the agent terminates no
closer to the goal than it began. We compute Equation~\ref{eq:progress} under
two cost functions. \emph{Tile progress} measures $d$ in grid cells and ignores
mechanisms. \emph{Action progress} measures $d$ in executable actions and
accounts for mechanisms. We report action progress throughout, because tile
progress rewards an agent for standing near a goal that is sealed behind a door
it never opened.

The oracle computes $d(s)$ by breadth-first search over the joint state of
position, facing direction, inventory, open doors, open gates, occupied tile,
and goal tile. Because mechanism state is part of the search state, $d(s)$
counts the turns, moves, pickups, and toggles needed to operate the remaining
mechanisms as well as the navigation steps, and it is conditioned on what the
agent has already achieved. Acquiring the correct key or opening a door
therefore lowers $d(s)$ and raises progress, so the metric credits subgoals
without any hand-specified subgoal list.

\subsection{Termination}

An episode ends in one of three ways: the agent reaches the goal, it exhausts a
step cap of three times the oracle-optimal action count for that instance, or a
stall watchdog fires. The watchdog ends an episode after 30 consecutive steps
with no change in position, inventory, or mechanism state. Setting the step cap
as a multiple of the oracle count makes the budget proportional to instance
difficulty rather than a flat allowance that is generous for small mazes and
punishing for large ones.

\subsection{Task pool and evaluation set}

We built a pool of 214 mazes by a combination of hand design and generation with
a language model, and validated every instance. The extensive validation process is detailed in Appendix \ref{sec:mazevalidation}. Oracle-optimal solutions in the
pool range from 9 to 106 executable actions.

From the validated pool we selected 50 mazes stratified across five bands of
oracle-optimal path length: 2 mazes at 20 to 25 actions, 12 at 26 to 39, 14 at
40 to 59, 8 at 60 to 79, and 14 above 80. Grid sizes are 11 mazes at
$8 \times 8$, 19 at $10 \times 10$, and 20 at $14 \times 14$. The set spans
three instance categories: \emph{scale} instances require navigation only,
\emph{mechanism} instances require navigation and mechanism operation, and
\emph{distractor} instances add decoys to both.

\section{Fixing the evaluation protocol}
\label{sec:protocol}

The protocol determines the parameter configurations of the environment and how a model interacts with the environment. This is a
confound if chosen after seeing benchmark scores. Reported performance is
sensitive to prompt formatting and to single-prompt evaluation
\citep{sclar2024,mizrahi2024}, so we fixed the protocol first, on data we do not
report scores on. We used 15 mazes held out from the 50 evaluation mazes. The
sweep holds a baseline fixed and varies 12 conditions across seven axes, each
condition changing exactly one field, for 540 episodes in total
(Table~\ref{tab:protocol}). Condition ordering is stable across the three
models (Figure~\ref{fig:protocol-heatmap}, Appendix~\ref{app:figs}), which is
what allows one protocol to be fixed for all of them.

The baseline gives the model the task, the egocentric action space, and the
positions of the mechanisms, but nothing about how any mechanism operates. The
current state is supplied as an image with a text description of it, preceded by
images of the three previous states. The model is asked for one next action,
each query is a self-contained message with no accumulated turn history, and one
in-context example of a completed maze is supplied \citep{brown2020}.

\begin{table}[htbp]
    \caption{Protocol sweep. Twelve conditions across seven axes, each changing
  exactly one field relative to \texttt{prompt/standard}. Solves pooled over
  three models, out of 45 episodes per condition.}
  \label{tab:protocol}
  \centering
  \small
  \begin{tabular}{p{2.6cm}p{8.0cm}>{\raggedleft\arraybackslash}p{1.2cm}}
    \toprule
    Condition & What it changes relative to the baseline & Solves \\
    \midrule
    \texttt{prompt/standard} & Baseline: task, action space, mechanism
      positions, no rules & 27 \\
    \texttt{prompt/minimal} & Reduced to task and action space & 27 \\
    \texttt{prompt/verbose} & Extended with explanations of each mechanism & 25 \\
    \texttt{icl/zero\_shot} & In-context example removed & 25 \\
    \texttt{qry/subgoal} & Plans several steps per turn instead of one action & 20 \\
    \texttt{qry/full\_traj} & Emits the entire action sequence in one turn & 11 \\
    \texttt{ctx/text\_summary} & Three previous frames replaced by a text
      summary & 17 \\
    \texttt{ctx/current} & History removed, current frame only & 14 \\
    \texttt{hist/multiturn} & Chat history accumulated over three turns & 15 \\
    \texttt{act/cardinal} & Egocentric directions replaced by cardinal
      directions & 12 \\
    \texttt{obs/image\_only} & Text description of the current state removed
      & \textbf{5} \\
    \texttt{baseline+thinking} & Extended thinking enabled (not one of the
      seven axes) & 30 \\
    \bottomrule
  \end{tabular}
\end{table}

Three results follow. Removing the text description of the current state is by
far the largest single effect, reducing pooled solves from 27 to 5. Prompt
content barely matters: standard, minimal, and verbose yield 27, 27, and 25
solves, and the verbose condition scores no higher than the baseline even though
it states how each mechanism operates. Performance falls monotonically as the
model commits further ahead, from 27 solves for one action per turn to 20 for
subgoal planning to 11 for a full trajectory, consistent with the value of
interleaving action and observation \citep{yao2023}.

We fixed the protocol as follows: extended thinking enabled for all models with
an equal 64k output token budget; the minimal prompt, since prompt content does
not separate the conditions and the minimal variant states the least; and the
current state supplied as an image with no accompanying text.

The last choice needs comment, because \texttt{obs/image\_only} is the weakest
condition in the sweep and we selected it deliberately rather than selecting the
setting that maximizes measured performance. Under image-only observation the
consequence of an action must be recovered by comparing the frame before it with
the frame after it, which is the inference we intend to measure and which
matches an interface that returns a rendered view rather than a structured diff.
The cost is that absolute scores in Section~\ref{sec:results} are lower than
they would be under the baseline protocol, and the two should not be compared
(Section~\ref{sec:limitations}).

\section{Results}
\label{sec:results}

This section reports the three models of the main evaluation, and Section~\ref{sec:frontier} extends it to two newer frontier models. The models and their configurations are described in Appendix~\ref{sec:modelconfig}.

\subsection{Solves}
\label{sec:solve-rate}

The three models solve 6 of 150 episodes.\footnote{Intervals on mean action
progress are 95\% bootstrap intervals over episodes \citep{efron1979}, resampled
by maze so that the three episodes on a given maze form a cluster
\citep{cameron2015}. We fit no regression to the solve outcome: with 6 positive
outcomes in 150 episodes the events per predictor fall far below an informative
range \citep{peduzzi1996}, so every regression below uses action progress as the
response.} Claude solves 4 of its 50 mazes, and
Kimi and Qwen each solve 1. Of the 50 mazes, 45 are solved by no model.

\begin{figure}[htbp]
  \centering
  \includegraphics[width=0.9\linewidth]{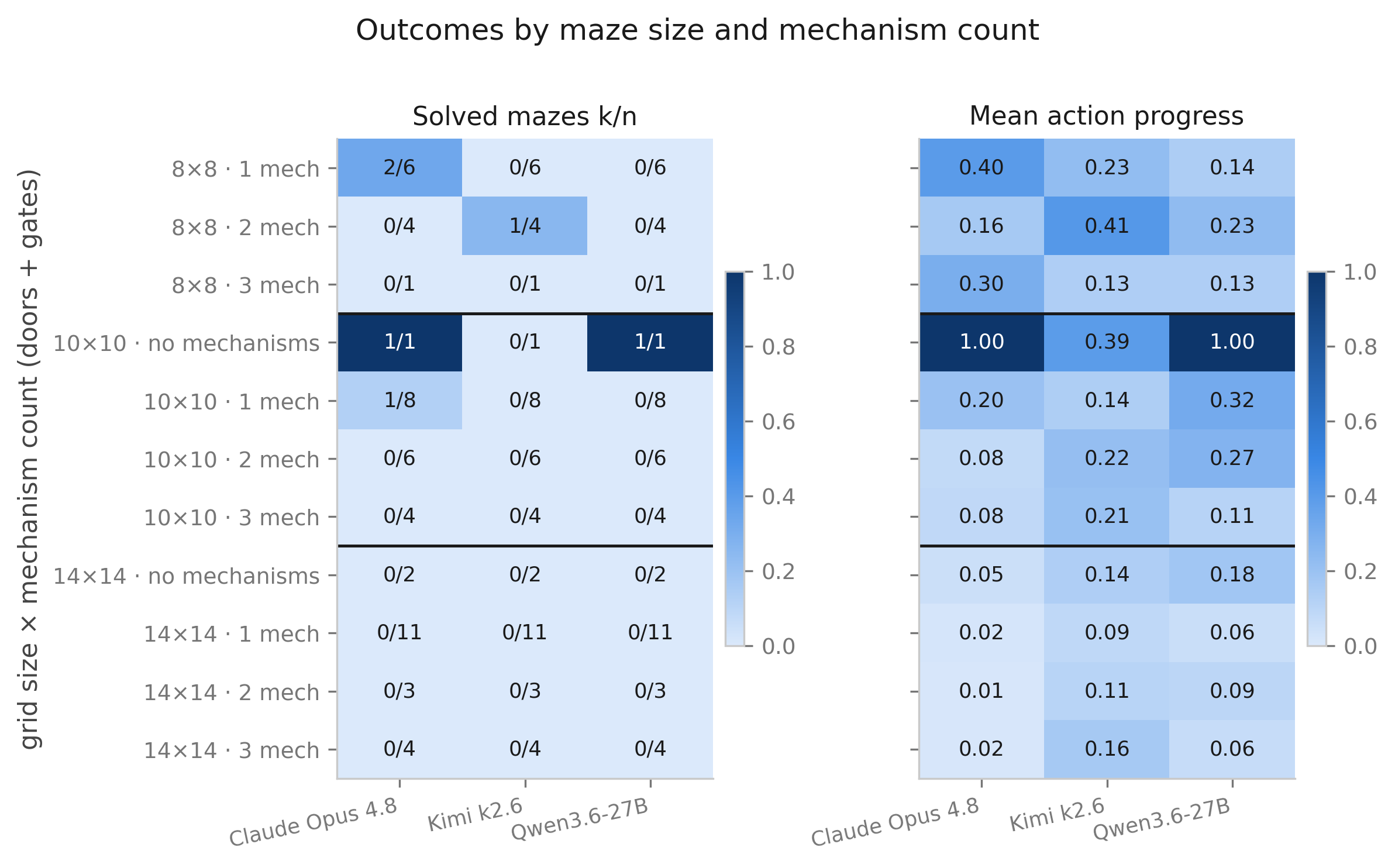}
  \caption{Solves by instance category for each model (left) and mean action
  progress by instance category for each model (right).}
  \label{fig:maze-type}
\end{figure}

Every solved instance shares three properties: it contains no switch and gate,
it contains no distractor, and its oracle-optimal path is 45 actions or fewer.
Solves by grid size are 3 of 33 episodes at $8 \times 8$, 3 of 57 at
$10 \times 10$, and 0 of 60 at $14 \times 14$
(Figure~\ref{fig:maze-type}).

\subsection{Progress}

Because 144 of 150 episodes end without a solve, action progress carries most of
the signal (Figure~\ref{fig:progress-grid}, Appendix~\ref{app:figs}).  The median closest approach across all episodes is 47.5 executable actions from the
goal, and the only episodes terminating within 3 actions of the goal are the 6
solves. The stall watchdog ended 139 of the 150 episodes, so agents spent most episodes in states
from which they made no measurable change rather than exhausting a search budget.

\subsection{What makes an instance hard}
\label{sec:difficulty}

Several declared parameters could govern difficulty: oracle-optimal path length,
grid size, the number and ordering of mechanisms, and the number of distractors.
For each we fit one linear regression predicting an episode's action progress
from that parameter and from an indicator for which model ran the episode, and
report $R^2$, the share of variance in action progress the fit accounts for.

\begin{figure}[htbp]
  \centering
  \includegraphics[width=0.9\linewidth]{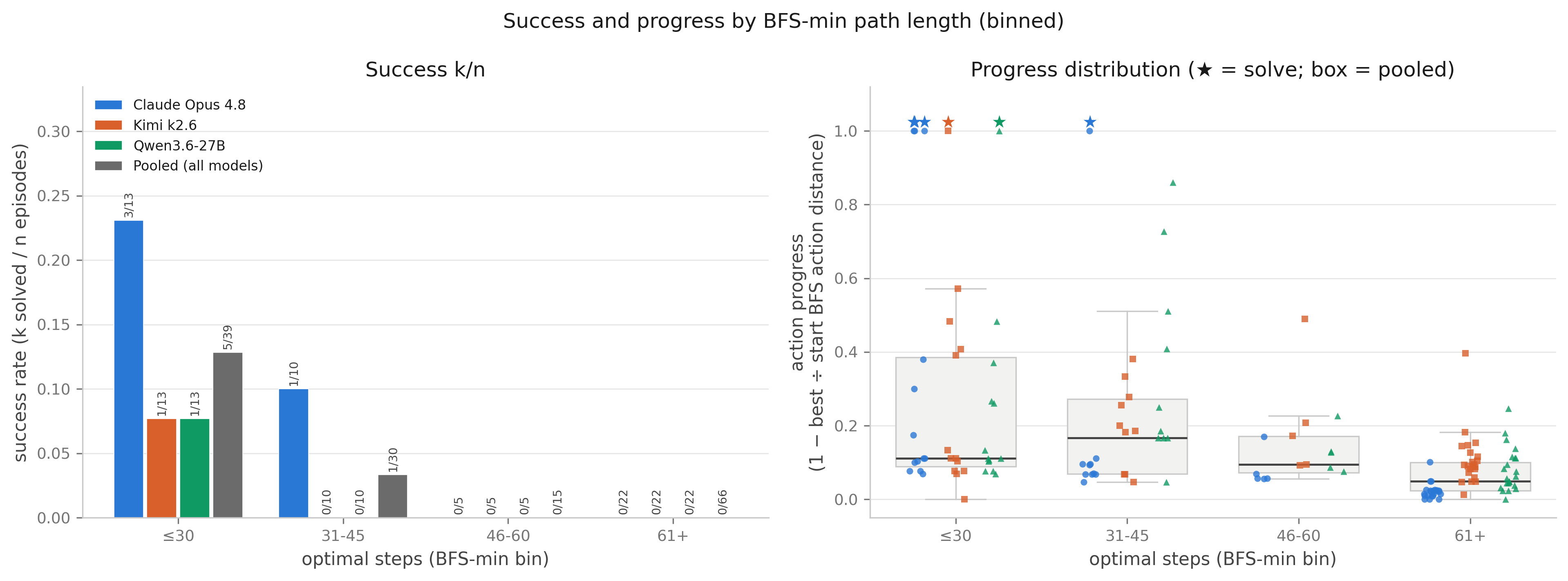}
  \caption{Solve rate by band of oracle-optimal path length (left) and mean
  action progress by the same bands (right). Both decline monotonically with
  path length.}
  \label{fig:path-length}
\end{figure}

Over the full set, path length dominates at $R^2 = 0.171$, ahead of gate count
at $0.122$, switch count at $0.104$, and distractor count at $0.030$. Grouped by path length, instances of 30 actions
or fewer produced 5 solves in 39 episodes at mean action progress $0.287$ (95\%
CI $0.163$ to $0.423$), 31 to 45 actions produced 1 in 30 at $0.240$ ($0.164$ to
$0.333$), 46 to 60 produced 0 in 15 at $0.141$ ($0.089$ to $0.205$), and 61 or
more produced 0 in 66 at $0.070$ ($0.058$ to $0.084$) (Figure~\ref{fig:path-length}).

The ranking is an artifact of the range over which each parameter varies. On the
69 episodes whose instances need 45 actions or fewer the ordering inverts: gate
count rises from $R^2 = 0.122$ to $0.255$ and switch count from $0.104$ to
$0.247$, while path length falls from $0.171$ to $0.012$. Path length binds
across the full set only because it varies over a range wide enough to drive
progress to near zero on its own; once the horizon is short enough for an agent
to plausibly reach the goal, mechanisms dominate. The two parameters therefore
bind in different regimes, which is a useful property for a benchmark that must
stay unsaturated.

\subsection{Attributing failure}
\label{sec:failure}

\paragraph{Switches and gates are not discovered.}
No solve occurred in any of the 105 episodes containing a switch and gate.
Across those episodes the models issued 132 toggle actions, of which 2 flipped a
switch, 15 opened a door, and 115 changed nothing. The models were not blocked
by navigation: an agent stood on a live switch in 38 of the 105 episodes, so it
reached the cell from which a single toggle would have opened the gate and did
not connect the two.

\paragraph{Keys and doors are handled with partial success.}
The same models produced 15 door openings, and 4 of the 6 solves came on
key-and-door instances. One Kimi solve required a key, a door, a second key, and
a second door in sequence, so multi-step ordering is within reach when the
semantics of each step are already familiar. The two families are similar in
interaction cost. They differ
in whether the association is available from pretraining: a key rendered in the
same colour as its door supplies the association visually, whereas a switch
rendered as a circle on a tile supplies nothing and its relationship to a
distant gate must be established by experiment. The gap between 15 door openings
and 2 switch flips is therefore attributable to affordance discovery rather than
to perception, navigation, or action execution, which the key-and-door results
show are sufficient here.

\paragraph{Interaction verbs are issued far more often than they apply.}
Across all episodes the models issued 220 pickup actions producing 27 key
acquisitions, and 132 toggle actions producing 2 switch flips and 15 door
openings, so roughly seven of every eight interaction attempts had no effect.
The models reach for interaction verbs far more often than the current state
affords an interaction, which indicates they are not tracking which affordances
are available from the cell they occupy.

\paragraph{Terminal behaviors differ by model.}
Beyond the shared switch failure, each model degrades in its own way
(Figure~\ref{fig:failure-modes}, Appendix~\ref{app:figs}). Claude fails
early relative to the other two, and in 91\% of the last 30 actions before a
failed episode it drives forward into a wall; in 41 of its 46 unsolved episodes,
wall collisions account for more than half of the final 30 actions, and the
collision share rises with step count as the share of successful moves falls.
Kimi retreads ground it has already visited, with revisits climbing from 2.7\%
of actions in the first quarter of an episode to 21.2\% in the last. Qwen
accumulates turns, whose share of its actions rises from 44.5\% to a peak of
59.6\% in the third quarter, while the share of steps that reach an unseen tile
collapses from 24.5\% to 2.4\%. These three patterns point to three different engineering problems, namely
collision feedback, visitation memory \citep{bellemare2016}, and exploration
policy \citep{pathak2017}, all of which a single headline score would have
reported identically as a zero.

\subsection{Test-time compute}

Kimi and Qwen spend more than 20 times as many output tokens per newly
discovered tile as Claude (Table~\ref{tab:compute}, Appendix ~\ref{sec:modelconfig}; Figure~\ref{fig:tokens-per-tile},
Appendix~\ref{app:figs}), and both cover
more of the maze than Claude does, yet each solves 1 episode against Claude's 4.
Additional reasoning under this protocol yields wider exploration without
yielding more solves. The protocol sweep points the same way from the other direction:
enabling extended thinking raised pooled solves on the held-out mazes from 27 to
30 of 45, a change small relative to the 22-solve effect of removing the state
description. We do not claim that test-time compute is unhelpful in general
\citep{snell2024}. We claim that in this environment it is not the binding
constraint, because the missing step is an inference about what a switch does,
and more tokens spent on a state that never included a flipped switch cannot
supply it.

Claude's thinking allocation moves in the opposite direction to what the task
requires. Its median thinking length after a move that changed nothing is 234
tokens across 1301 turns, against 579 tokens across 299 turns after a move that
made progress. Claude therefore reasons least at the moments when its
representation of the environment has just been contradicted.

\section{Extension to new frontier models}
\label{sec:frontier}

We extended our work to assess whether the attribution we detailed in Section~\ref{sec:results} holds for
stronger systems that are state-of-the-art. We evaluated GPT-6 Astra and Claude Opus 5.5 on the same 50
mazes under the protocol fixed in Section~\ref{sec:protocol}, with thinking at
the \texttt{xhigh} setting and the same 64k output token budget.
Table~\ref{tab:frontier} places them beside the three models of
Section~\ref{sec:results}.

\begin{table}[htbp]
\caption{Five models on the 50-maze evaluation set. Blocked moves are the
share of move-forward actions that hit a wall. Tokens per new tile are output
tokens per newly visited tile, pooled over episodes.} 
\label{tab:frontier}
\centering
\small
\begin{tabular}{lrrrr}
  \toprule
  Model & Solves & Mean action progress & Blocked moves & Tokens per new tile \\
  \midrule
  Claude Opus 4.8 & 4/50  & 0.14 & 82\% & 6{,}297 \\
  Kimi K2.6       & 1/50  & 0.18 & 52\% & $\geq$162{,}222 \\
  Qwen 3.6-27B    & 1/50  & 0.18 & 51\% & 123{,}120 \\
  \midrule
  Claude Opus 5.5 & 7/50  & 0.34 & 2\%  & 4{,}504 \\
  GPT-6 Astra     & 11/50 & 0.45 & 7\%  & 2{,}172 \\
  \bottomrule     
\end{tabular}
\end{table}

\paragraph{Horizon length is no longer a limiting factor.}
Astra solves 11 mazes and Opus 5.5 solves 7, against at most 4 for any earlier
model, and 38 of the 50 mazes remain unsolved by all five models. Astra solves
three $14 \times 14$ mazes at 77 to 85 actions and Opus 5.5 solves four at 77
to 89 actions, where the earlier models solved 0 of 60 episodes. Path length
accounts for $R^2 = 0.019$ of the variance in action progress for Astra and
$0.003$ for Opus 5.5, against $0.212$ for Claude Opus 4.8, $0.179$ for Qwen,
and $0.109$ for Kimi. This is the regime Section~\ref{sec:difficulty}
anticipates: once the horizon is within reach, mechanisms dominate.

\paragraph{The switch-gate mechanism failure persists.}
Every solve by either model is on one of the 15 mazes without a switch. Astra
solves 11 of them and Opus 5.5 solves 7, and neither solves any of the 35
mazes with a switch and gate. Holding path length fixed, a switch and gate
lowers action progress by $0.57$ for Astra and $0.38$ for Opus 5.5, against
$0.07$ to $0.22$ for the earlier models, so the gap widens as overall
performance rises. Both models fail by applying a familiar convention to an
unfamiliar object. Astra treats the switch as an item, issuing pickup on it 63
times, aims its 66 no-effect toggles at the gate as if it were a door, and
never toggles a switch while standing on it. Opus 5.5 aims 108 pickups and one
toggle at switches, and 199 of its 200 no-effect toggles at a gate from an 
adjacent cell.

\paragraph{Terminal behaviors are still distinct across models.}
All 39 of Astra's failures and all 43 of Opus 5.5's failures end in a stall. Astra hits a wall on 5\% of moves in its final
30 steps, and Opus 5.5 hits a wall once in 1{,}065 forward moves. Astra loops
over explored ground. Its failed episodes reach a median of 16 tiles, against 4
to 7 for the earlier models, but its final 30 steps cover a median of 7 cells,
all previously visited, which is the visitation-memory failure of Kimi in
Section~\ref{sec:failure} at a larger scale. Opus 5.5 repeats actions that have
no effect: 59\% of its final 30 actions before a stall change nothing, and 472
of its 2{,}599 prompts are identical to the one before. Most of these are
pickups aimed at a key from the adjacent cell, 570 against 7 issued while
standing on one, which follows the convention of standard MiniGrid
\citep{chevalier2023} rather than of this environment, and all 8 of its
failures on switchless mazes end in this loop.

\section{Discussion}

\paragraph{A two-dimensional maze is not an easy task for current models.}
The substrate is deliberately simple and the models are among the most capable
available, yet 38 of 50 instances went unsolved by all five models, and raising one parameter is
enough to drive action progress from $0.287$ to $0.070$ without touching any
mechanism. Since we regard this substrate as a heavily simplified stand-in for
workflows that require many actions in an interface whose conventions are not
given, the results indicate headroom on long-horizon reliability rather than a
difficulty peculiar to mazes.

\paragraph{Discovery through interaction is the sharpest gap.}
The switch and gate result is the strongest finding here because the comparison
is controlled: both mechanisms need one toggle from one cell, both sit in the
same renderer and action space, and the models reached the switch cell in 38
episodes without exploiting it. What separates them is whether the association
can be read off the rendering using prior knowledge. Agents deployed against
real interfaces meet controls whose function is not stated, and establishing
function by experiment is what these models did not do. Evidence that
vision-language models resolve fine-grained visual relations poorly
\citep{fu2024,rahmanzadeh2024} suggests part of this gap may be perceptual
rather than inferential, which the text-domain projection we discuss in Section \ref{sec:conclusion} is designed to
separate.

\paragraph{Contamination and saturation.}
Instances are compositions of declared parameters, so difficulty can be raised
along any axis as models improve, which is the property dynamic benchmarks were
introduced to obtain \citep{kiela2021}. Because instances are newly authored
rather than drawn from an existing corpus, and because the oracle is computed
rather than annotated, there is no answer key to memorize.

\section{Limitations}
\label{sec:limitations}

\paragraph{The benchmark does not rank these models.}
With 6 solves in 150 episodes, and with the ranking by action progress inverting
the ranking by solves, this version of the benchmark does not separate the
capabilities of frontier models. Its use at present is to localize failures
rather than to rank systems. The two newer models in Section~\ref{sec:frontier} separate from the earlier three on both solves and action progress, which suggests the benchmark begins to rank systems once solve rates rise.

\paragraph{Absolute scores are protocol-dependent by construction.}
We deliberately fixed the observation modality to the weakest condition in the
sweep (Section~\ref{sec:protocol}). Under the baseline protocol the same models
recorded 27 pooled solves of 45 on held-out mazes. The near-zero headline
numbers should therefore be read as performance under image-only observation,
not as a protocol-independent capability estimate.

\paragraph{The discovery finding rests on one mechanism.}
Our conclusion about affordance discovery rests on the contrast between one
unfamiliar mechanism and one familiar one. A second and third unfamiliar
mechanism with different visual forms and interaction costs would substantially
strengthen it, and would rule out the possibility that the specific rendering of
a switch, rather than its unfamiliarity, drives the result.

\paragraph{Visual history is capped at three frames.}
To keep the harness minimal and input costs bounded, we supply three previous
frames. We did not measure how performance changes as that window grows, so we
cannot separate a failure of inference from a failure of context.

\paragraph{No human baseline.}
We did not run a human baseline, so we cannot state the size of the gap between
these models and human performance on the same instances, nor confirm that the
instances are straightforward for people.

\paragraph{The agent is not told its remaining budget.}
Models are not informed how many steps remain. We did not ablate this, so we
cannot say whether the wandering behavior in Section~\ref{sec:failure} reflects
an absence of budget awareness.

\section{Conclusion and future work}
\label{sec:conclusion}

We built a maze environment in which mechanisms can be varied one at a time
against a fixed renderer, action space, and scoring function, and in which an
exact oracle over the joint environment state supplies per-instance difficulty
and graded partial credit. Three vision-language models solved 6 of 150
episodes, attributed to horizon length, affordance discovery, and three distinct
per-model terminal behaviors. Two newer frontier models solved 18 of 100 episodes and depended far less on horizon length, but neither solved a maze containing a switch, so affordance discovery remains the main limit as capability rises. Three extensions follow directly: unfamiliar mechanisms with different visual forms, to test whether the discovery finding generalizes; a text-only rebuild of the same instances, to separate perception from planning and execution; and a projection into three-dimensional simulation, to measure transfer of a solved task into a new modality and action space.

\begin{ack}
We thank Victor Barres for feedback on the design of this environment and benchmark, and for his insights on the evaluation space more broadly. We also thank Greg Kamradt and Yuansheng Ni for early comments on the direction and framing of this work.

This work was funded by Fig, which also covered the model API and compute costs of all evaluations reported in this paper.
\end{ack}

\bibliographystyle{plainnat}
\bibliography{references}

\clearpage
\appendix

\section{Mechanism details}
\label{app:mechanisms}

\begin{table}[htbp]
\label{table:mechanisms}
  \caption{Each mechanism family isolates one requirement of the task. Because
  families are independent, adding or removing one changes the requirement it
  isolates while leaving the renderer, the action space, and the scoring
  function unchanged.}
  \label{tab:mechanisms}
  \centering
  \begin{tabular}{p{5.4cm}p{7.4cm}}
    \toprule
    Mechanism family & Requirement isolated \\
    \midrule
    Key and door & Operating a mechanism whose semantics are available from
      pretraining \\
    Switch and gate & Discovering an affordance whose semantics are not
      available from pretraining \\
    Dependency chain & Reasoning about order of operations \\
    Distractors: wrong key, inactive switch, dead-end branch &
      Recovering from a wasted excursion \\
    \bottomrule
  \end{tabular}
\end{table}

\section{Additional figures}
\label{app:figs}

\begin{figure}[htbp]
  \centering
  \includegraphics[width=0.9\linewidth]{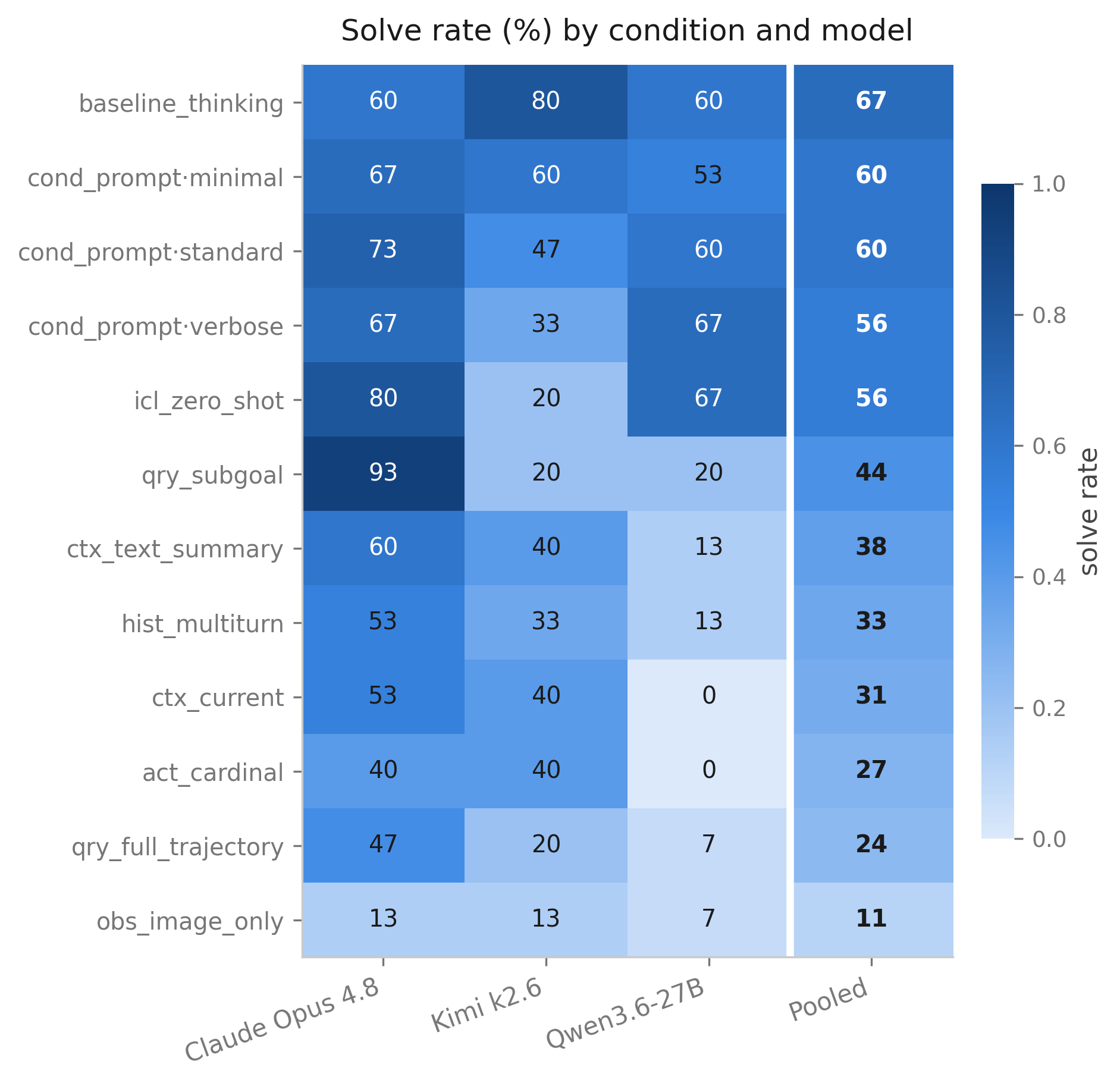}
  \caption{Solves per model under each protocol condition on the 15 held-out
  mazes. The ordering of conditions is stable across models, which is what
  allows a single protocol to be fixed for all three.}
  \label{fig:protocol-heatmap}
\end{figure}

\begin{figure}[htbp]
  \centering
  \includegraphics[width=0.49\linewidth]{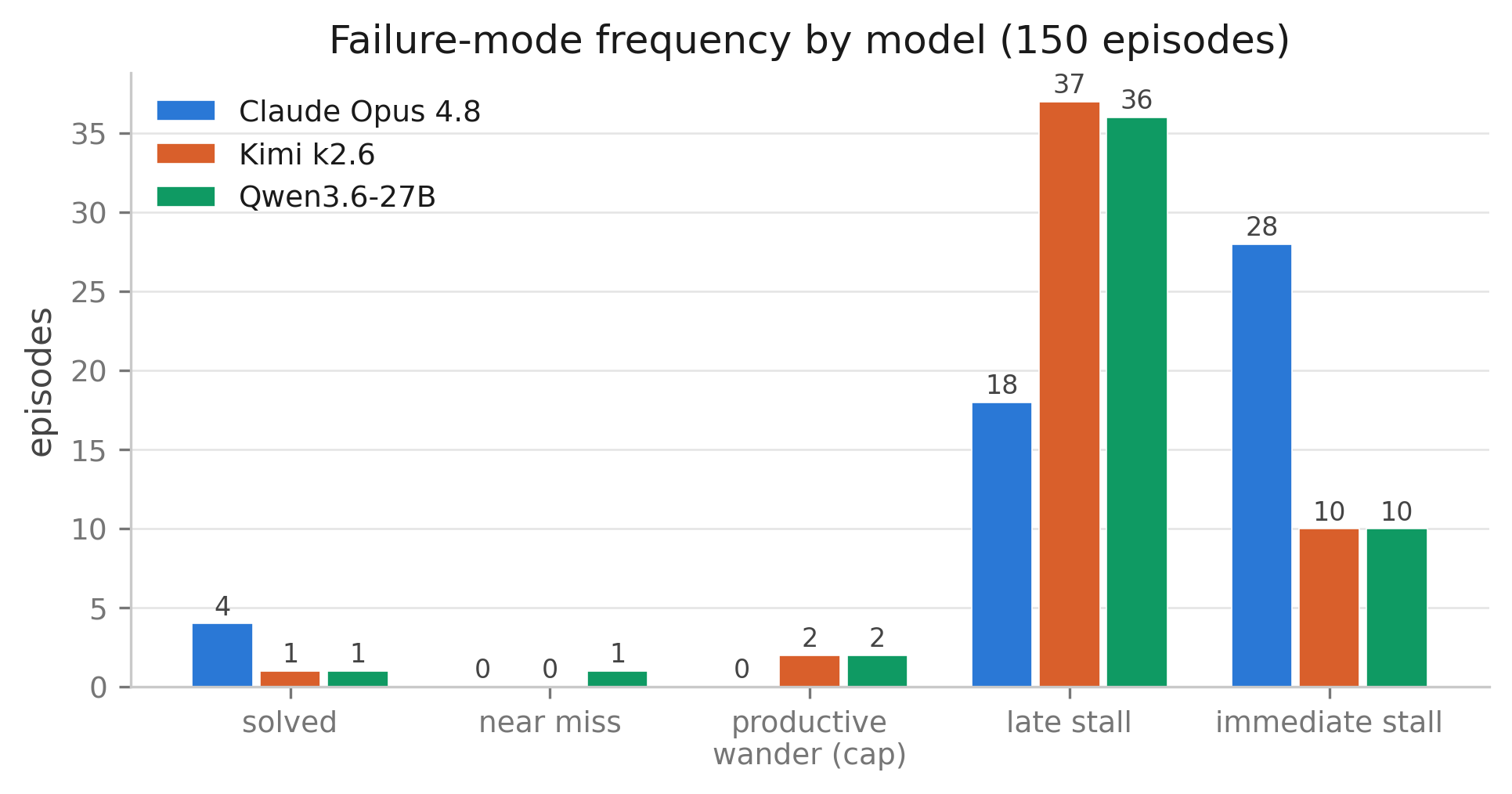}\hfill
  \includegraphics[width=0.49\linewidth]{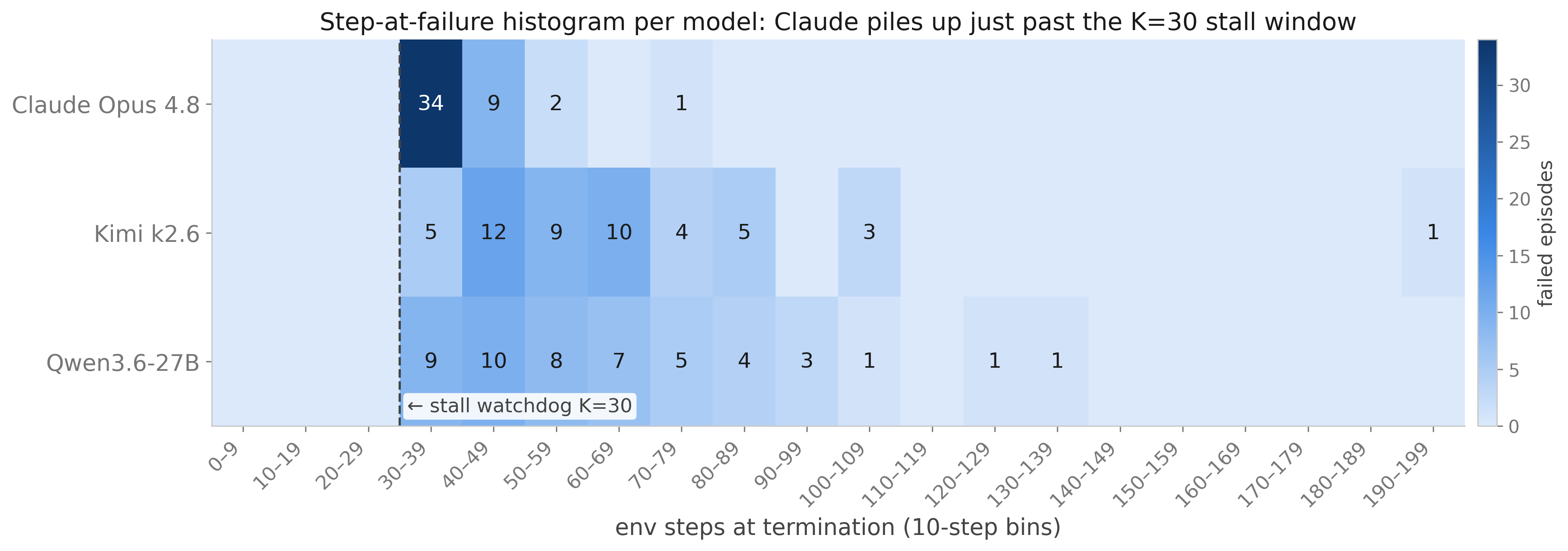}
    \caption{Failure-mode frequency by model (left), distribution of the
  terminal step by model (right).}
  \label{fig:failure-modes}
\end{figure}

\begin{figure}[htbp]
  \centering
  \includegraphics[width=0.86\linewidth]{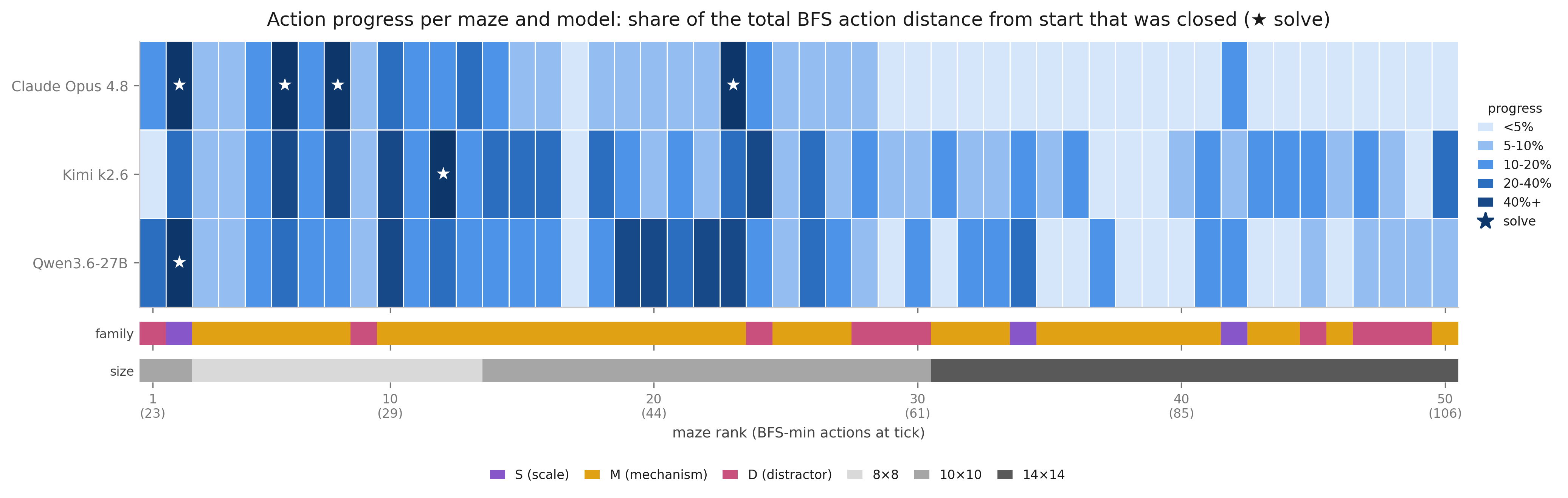}
  \caption{Action progress per model and maze. A star marks a solve. Action
  progress is one minus the remaining oracle actions at the terminal state
  divided by the oracle actions from the start state, per
  Equation~\ref{eq:progress}.}
  \label{fig:progress-grid}
\end{figure}

\begin{figure}[htbp]
  \centering
  \includegraphics[width=0.6\linewidth]{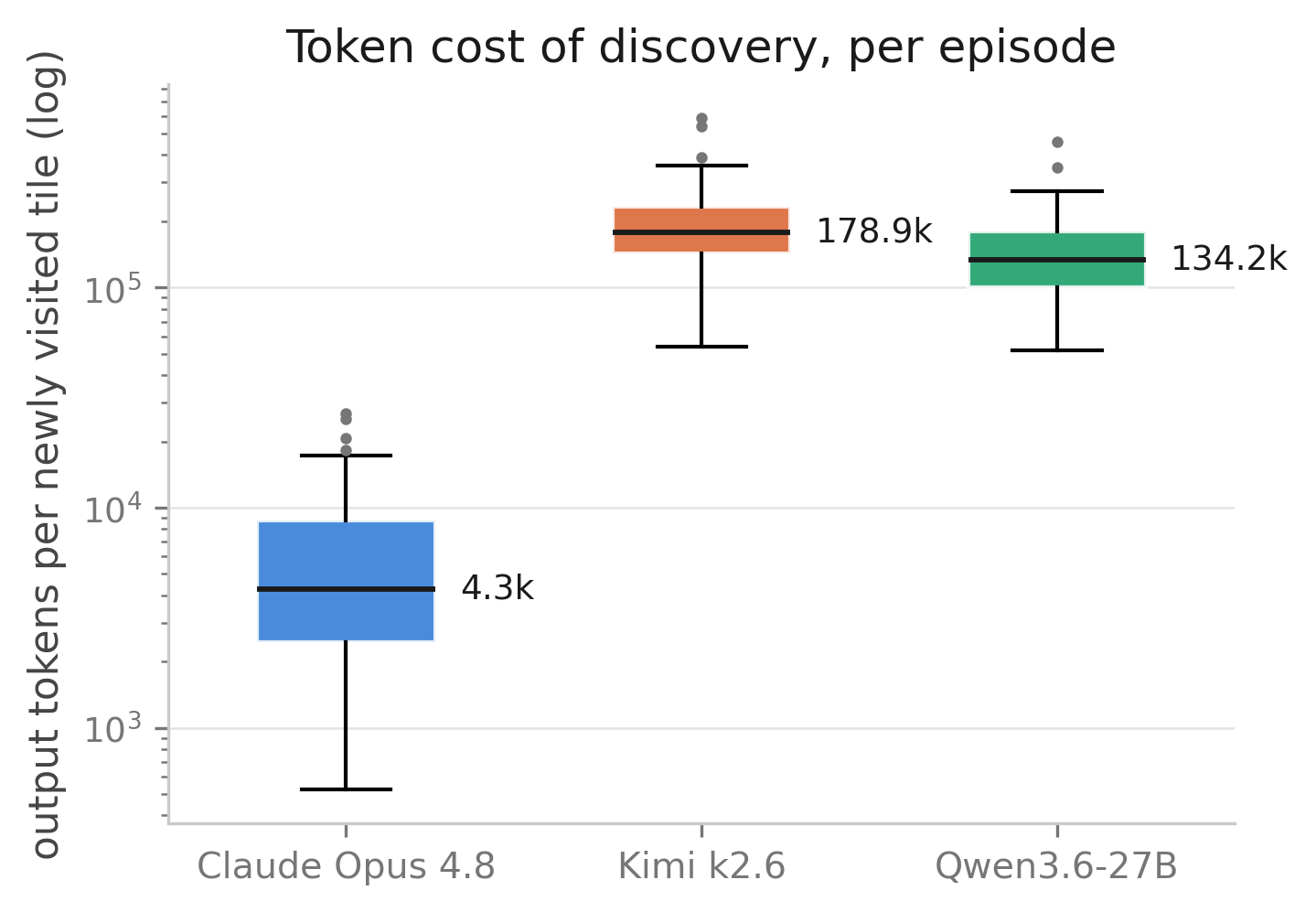}
  \caption{Output tokens per newly discovered tile per episode, shown as a
  boxplot on a log axis with medians labelled.}
  \label{fig:tokens-per-tile}
\end{figure}

\section{Maze validation}
\label{sec:mazevalidation}

Validation of each maze had four stages. Reachability was confirmed by exhaustive search
under a cap of 500{,}000 states, and the optimal action count was computed with
the oracle of Section~\ref{sec:scoring}. A mechanism necessity check flags any
mechanism whose removal leaves the instance solvable, since such a mechanism is
decorative. A chain ordering check confirms that each mechanism in a dependency
chain is unreachable until its predecessor has been operated. A distractor
safety check confirms that no single distractor interaction can render the
instance unsolvable, so a wasted excursion is recoverable by construction and a
failure to recover is attributable to the agent.

\section{Models and run configuration}
\label{sec:modelconfig}

We evaluate Claude Opus 4.8, Kimi K2.6, and Qwen 3.6-27B. Claude was run with
adaptive thinking at the \texttt{xhigh} setting, and Kimi and Qwen were run with
their thinking modes enabled. All three received an equal budget of 64k output
tokens per query. Claude and Kimi were queried through their native APIs. Qwen
was served locally without quantization on three A100 80GB accelerators in parallel. GPT-6 Astra and Claude Opus 5.5 (Section~\ref{sec:frontier}) were run with thinking at the \texttt{xhigh} setting and the same 64k output token budget per query.

\begin{table}[htbp]
  \caption{Token statistics and cost per model.}
  \label{tab:compute}
  \centering
  \small
  \begin{tabular}{p{5.0cm}rrr}
    \toprule
    & Claude Opus 4.8 & Kimi K2.6 & Qwen 3.6-27B \\
    \midrule
    Median output tokens per query & 289 & 20{,}184 & 16{,}444 \\
    Output tokens, total (pooled)  & 2.06M & $\geq$67.0M & 48.0M \\
    Tokens per new tile (pooled)   & 6{,}297 & $\geq$162{,}222 & 123{,}120 \\
    \midrule
    Observed cost & \$43.64 & \$264.52 & ${\sim}165$ A100 hrs \\
    \bottomrule
  \end{tabular}
\end{table}

\section{Environment and reproduction details}
\label{app:release}

The environment, the evaluation harness, the run infrastructure, and the oracle
implementation are released under the MIT licence at the repository given in footnote~\ref{fn:code}. The release also includes prompts for all 12 protocol conditions, and the per-episode logs for all 790 episodes reported in this paper.

\newpage
\section*{NeurIPS Paper Checklist}

\begin{enumerate}

\item {\bf Claims}
    \item[] Question: Do the main claims made in the abstract and introduction accurately reflect the paper's contributions and scope?
    \item[] Answer: \answerYes{}
    \item[] Justification: The environment, the oracle, the protocol sweep, and the evaluation claimed in the abstract and introduction are established in Sections~\ref{sec:env}, \ref{sec:protocol}, \ref{sec:results}, and \ref{sec:frontier}. Section~\ref{sec:limitations} bounds their scope: the benchmark localizes failures rather than ranking systems, and absolute scores hold only under the image-only protocol.

\item {\bf Limitations}
    \item[] Question: Does the paper discuss the limitations of the work performed by the authors?
    \item[] Answer: \answerYes{}
    \item[] Justification: Section~\ref{sec:limitations} covers six: no separation of frontier model capabilities at this solve rate, protocol-dependent absolute scores, a single unfamiliar mechanism behind the affordance-discovery finding, an unablated three-frame history cap, no human baseline, and no test of budget awareness.

\item {\bf Theory assumptions and proofs}
    \item[] Question: For each theoretical result, does the paper provide the full set of assumptions and a complete (and correct) proof?
    \item[] Answer: \answerNA{}
    \item[] Justification: The paper contains no theorems. Equation~\ref{eq:progress} defines a metric, and the oracle's optimality holds by construction as an exhaustive search over a finite deterministic state space.

    \item {\bf Experimental result reproducibility}
    \item[] Question: Does the paper fully disclose all the information needed to reproduce the main experimental results of the paper to the extent that it affects the main claims and/or conclusions of the paper (regardless of whether the code and data are provided or not)?
    \item[] Answer: \answerYes{}
     \item[] Justification: Sections~\ref{sec:env} and \ref{sec:protocol} specify the action space, mechanism semantics, observation format, scoring, termination rules, and all twelve protocol conditions. Appendices~\ref{sec:modelconfig}, \ref{sec:mazevalidation}, and \ref{app:release} give the model configurations, the instance validation procedure, and the released artifacts, which include the exact prompts and per-episode logs for all 790 episodes. We do not publicly release the mazes that are part of the evaluation set itself to avoid issues with respect to contamination.

\item {\bf Open access to data and code}
    \item[] Question: Does the paper provide open access to the data and code, with sufficient instructions to faithfully reproduce the main experimental results, as described in supplemental material?
    \item[] Answer: \answerYes{}
    \item[] Justification: The environment, harness, run infrastructure, and oracle are released under the MIT licence at the repository linked from the introduction. Appendix~\ref{app:release} lists the contents.

\item {\bf Experimental setting/details}
    \item[] Question: Does the paper specify all the training and test details (e.g., data splits, hyperparameters, how they were chosen, type of optimizer) necessary to understand the results?
    \item[] Answer: \answerYes{}
    \item[] Justification: No training is performed; Appendix~\ref{sec:modelconfig} gives the thinking settings and the equal 64k token budget, and Section~\ref{sec:env} gives the termination rules and the stratification of the 50-maze evaluation set. Section~\ref{sec:protocol} states that the protocol was fixed on a disjoint 15-maze sweep set before any evaluation number was measured, so it is not a parameter fitted to the reported results.

\item {\bf Experiment statistical significance}
    \item[] Question: Does the paper report error bars suitably and correctly defined or other appropriate information about the statistical significance of the experiments?
    \item[] Answer: \answerYes{}
    \item[] Justification: Mean action progress carries 95\% bootstrap intervals resampled by maze so that the three episodes on a maze form a cluster, as stated in the footnote in Section~\ref{sec:solve-rate}. Solve counts and $R^2$ values are point estimates, and we deliberately fit no regression to the binary solve outcome, since 6 positive outcomes in 150 episodes puts events per predictor below an informative range.

\item {\bf Experiments compute resources}
    \item[] Question: For each experiment, does the paper provide sufficient information on the computer resources (type of compute workers, memory, time of execution) needed to reproduce the experiments?
    \item[] Answer: \answerYes{}
    \item[] Justification: Appendix~\ref{sec:modelconfig} reports the 64k output
    token budget per query and the three A100 80GB accelerators serving the local
    model, and Table~\ref{tab:compute} reports per-model token totals and the
    cost of the evaluation.
    
\item {\bf Code of ethics}
    \item[] Question: Does the research conducted in the paper conform, in every respect, with the NeurIPS Code of Ethics \url{https://neurips.cc/public/EthicsGuidelines}?
    \item[] Answer: \answerYes{}
    \item[] Justification: The work involves no human subjects, no personal data, and no scraped content; all instances are synthetic and newly authored, and all evaluated models were accessed under their providers' terms.

\item {\bf Broader impacts}
    \item[] Question: Does the paper discuss both potential positive societal impacts and negative societal impacts of the work performed?
    \item[] Answer: \answerYes{}
    \item[] Justification: The positive impact is diagnostic: attributing a failure to a specific cause lets a practitioner choose an intervention instead of inferring one from an aggregate score, as Section~\ref{sec:failure} shows for three models whose headline scores are near-identical. The foreseeable risk is that near-zero scores measured under a deliberately hard protocol are quoted as a capability ranking, which Section~\ref{sec:limitations} rules out directly.
    
\item {\bf Safeguards}
    \item[] Question: Does the paper describe safeguards that have been put in place for responsible release of data or models that have a high risk for misuse (e.g., pre-trained language models, image generators, or scraped datasets)?
    \item[] Answer: \answerNA{}
    \item[] Justification: The release contains only synthetic maze instances, an oracle, an evaluation harness, prompts, and run logs. It includes no pretrained or generative models and no scraped or human-derived data.

\item {\bf Licenses for existing assets}
    \item[] Question: Are the creators or original owners of assets (e.g., code, data, models), used in the paper, properly credited and are the license and terms of use explicitly mentioned and properly respected?
    \item[] Answer: \answerYes{}
    \item[] Justification: MiniGrid \citep{chevalier2023} is cited in Section~\ref{sec:frontier} and used under its Apache-2.0 licence; our own release is licensed MIT. The evaluated models were accessed under their providers' terms (Appendix~\ref{sec:modelconfig}), and no existing datasets are used.

\item {\bf New assets}
    \item[] Question: Are new assets introduced in the paper well documented and is the documentation provided alongside the assets?
    \item[] Answer: \answerYes{}
    \item[] Justification: Appendix~\ref{app:release} documents what the release contains and Appendix~\ref{sec:mazevalidation} the four-stage validation every instance passed. The repository carries an MIT licence file and usage documentation.

\item {\bf Crowdsourcing and research with human subjects}
    \item[] Question: For crowdsourcing experiments and research with human subjects, does the paper include the full text of instructions given to participants and screenshots, if applicable, as well as details about compensation (if any)? 
    \item[] Answer: \answerNA{}
    \item[] Justification: The paper involves no crowdsourcing and no human subjects. Section~\ref{sec:limitations} notes that no human baseline was run and lists this as a limitation.

\item {\bf Institutional review board (IRB) approvals or equivalent for research with human subjects}
    \item[] Question: Does the paper describe potential risks incurred by study participants, whether such risks were disclosed to the subjects, and whether Institutional Review Board (IRB) approvals (or an equivalent approval/review based on the requirements of your country or institution) were obtained?
    \item[] Answer: \answerNA{}
    \item[] Justification: The research involves no human subjects or study participants, so no IRB approval or equivalent review was required.

\item {\bf Declaration of LLM usage}
    \item[] Question: Does the paper describe the usage of LLMs if it is an important, original, or non-standard component of the core methods in this research? Note that if the LLM is used only for writing, editing, or formatting purposes and does \emph{not} impact the core methodology, scientific rigor, or originality of the research, declaration is not required.
    \item[] Answer: \answerYes{}
    \item[] Justification: Language models generated part of the 214-instance pool alongside hand design, as stated in Section~\ref{sec:env}. Every instance had to pass the four-stage validation in Appendix~\ref{sec:mazevalidation} before entering the pool, so no reported property depends on the generating model being correct.

\end{enumerate}

\end{document}